\documentclass[conference]{IEEEtran}
\IEEEoverridecommandlockouts
\usepackage{cite}
\usepackage{amsmath,amssymb,amsfonts}
\usepackage{algorithmic}
\usepackage{graphicx}
\usepackage{textcomp}
\usepackage{xcolor}
\usepackage{booktabs}
\usepackage{tabularx}
\usepackage{subfigure}
\usepackage{stfloats}
\usepackage{float}
\usepackage{multirow}
\counterwithout{figure}{section}
\def\BibTeX{{\rm B\kern-.05em{\sc i\kern-.025em b}\kern-.08em
    T\kern-.1667em\lower.7ex\hbox{E}\kern-.125emX}}
\begin{document}

\title{Adaptive Activation Network for Low Resource Multilingual Speech Recognition\\
}

\author{
\IEEEauthorblockN{Jian Luo, Jianzong Wang*\thanks{*Corresponding author: Jianzong Wang, jzwang@188.com}, Ning Cheng, Zhenpeng Zheng, Jing Xiao}
\IEEEauthorblockA{\textit{Ping An Technology (Shenzhen) Co., Ltd.}}
\textit{Shenzhen, China}
}

\maketitle

\begin{abstract}
Low resource automatic speech recognition (ASR) is a useful but thorny task, since deep learning ASR models usually need huge amounts of training data. The existing models mostly established a bottleneck (BN) layer by pre-training on a large source language, and transferring to the low resource target language. In this work, we introduced an adaptive activation network to the upper layers of ASR model, and applied different activation functions to different languages. We also proposed two approaches to train the model: (1) cross-lingual learning, replacing the activation function from source language to target language, (2) multilingual learning, jointly training the Connectionist Temporal Classification (CTC) loss of each language and the relevance of different languages. Our experiments on IARPA Babel datasets demonstrated that our approaches outperform the from-scratch training and traditional bottleneck feature based methods. In addition, combining the cross-lingual learning and multilingual learning together could further improve the performance of multilingual speech recognition.
\end{abstract}

\begin{IEEEkeywords}
Adaptive Activation Network, Multilingual, Low Resource Speech Recognition
\end{IEEEkeywords}

\section{Introduction}
Recently, end-to-end Automatic Speech Recognition (ASR) has attracted a lot of attention due to the significant performance improvement brought by deep neural networks~\cite{joint1, Hybrid2,Luo2021Multi}. However, building an ASR system usually requires a large amount of transcripted training data. It is known that labeling speech data requires huge labor resources. In addition, compared with rich-resource language, the dataset of low-resource language is more difficult to obtain~\cite{am}. Hence, it has great significance to establish an efficient and high-accuracy low resource language, as well as multilingual speech recognition system~\cite{2017An, icassp20201,icassp20191}.

Some works have been devoted to the development of low-resource ASR methods~\cite{Jia2020Large,icassp20202}. The most common method applied to low resource tasks is transfer learning~\cite{Luo2021Cross}. These methods use a large-scale corpus to pre-train the model, and utilize this pre-trained model to extract bottleneck (BN) feature layers for the target language~\cite{alter, ACL1}. Then, the upper layers of the target language model are fine-tuned on the target corpus. The bottom layers might contain the common speech features, such as phoneme, spectrum, intensity, etc. In this work, we used an end-to-end CNN-RNN-DNN (Convolutional-Recurrent-Deep) ASR network, which is inspired by DeepSpeech2~\cite{ds2}. The bottom convolutional layers are designed to extract local speech features, and the upper recurrent and deep neural layers can model different language information on different target languages.

Multi-task learning is another effective method for multilingual low resource ASR systems~\cite{icasspx}. Multilingual ASR establishes multi-task learning to train all of the target languages together. The typical structures are shared-hidden-layer (SHL) model~\cite{2016Multilingual, 2019Multilingual}, stacked-shared-exclusive (SSE) model~\cite{im}, and parallel-shared-exclusive (PSE) model~\cite{icasp1,yi2019Language}. Multilingual ASR systems usually include shared units and exclusive units. The shared units are leveraged to learn common features of each language, while the exclusive units are designed to train the monolingual target. However, the choice of shared and exclusive units is a manual and complicated process. Some researchers have designed a special structure to distinguish between the shared and exclusive units. In this work, we introduced an adaptive activation network to find suitable structure among different languages automatically.

Task Adaptive Activation Network (TAAN) is a flexible and accurate architecture for multi-task learning~\cite{2020Adaptive}. TAAN has proved to be an effective model for multi-domain video classification task. Because all the tasks share the same weight and bias parameters of the neural network, the complexity of TAAN is similar to that of a single task model. TAAN discovers the optimal knowledge sharing structure automatically through adjusting the task-adaptive activation functions according to the properties and quantities of each individual task. With its flexible structure and capability of dealing with task-imbalance problem, we find its great potential in ASR area. Thus, we introduced the adaptive activation network to the low resource multilingual ASR in this paper. We used the adaptive activation to replace the traditional activation function of RNN and DNN layers. The adaptive activation can automatically explore the language relationship by learning multiple language-adaptive activation functions. We also proposed two learning strategies to realize this language-adaptive characteristic. One is cross-lingual learning, replacing the language-adaptive activation from source to target language. The other is multilingual learning, training all of the target languages together with CTC loss of each language and relevant trace-norm loss among different languages.

In summary, the main contributions of this paper are as following:
\begin{itemize}
	\item Introduce adaptive activation network to low resource speech recognition, applying different activation functions to different languages.
	\item Propose a cross-lingual learning approach, replacing the activation function of upper layers for target language.
	\item Propose a multilingual learning approach, jointly training the CTC loss of each language and the relevance of different languages by trace-norm function.
	\item Combine the cross-lingual learning and multilingual learning together, further improving the performance of the multilingual speech recognition.
\end{itemize} 

\section{Related Works}

It has been shown that the performance of end-to-end ASR models degrades sharply on low-resource languages. Some researchers employed unsupervised or semi-supervised methods to exploit unlabeled data to alleviate the need for labeled data~\cite{Luo2021Dropout}. XLST is a cross-lingual self-training method, which is trained by maximizing the similarity between the target and source language~\cite{zhang2021xlst}. However, these methods still require massive unlabeled low-resource data. To solve this issue, some methods used transfer learning to improve the performance of low-resource languages by exploiting other source languages. \cite{yi2019Language} proposed to use an additional language discriminator in the adversarial SHL model, making the shared layers of source model can learn more language invariant features. \cite{icasp1} attempted to learn a language-invariant BN features through an attention based adversarial language identification. \cite{Kannan2019Large} found that combination of a language vector and language-specific adapter layers could handle the imbalance of training data across languages. \cite{Kumar2021AnEO} combined semi-supervised training and language adversarial transfer learning to improve the performance of Hindi ASR system in limited resource conditions. 

Multi-task learning has achieved great success in speech tasks. Some works established a multilingual speech recognition system by multi-task learning. Dalmia \textit{et al.}~\cite{dalmia2018sequence} used multi-task learning to train a multi-language model and then transfers the model to a specific language. Hou \textit{et al.}~\cite{hou2020large} proposed a large-scale multilingual transfer learning ASR, proving that pre-training could improve the performance of the model significantly on low-resource languages. \cite{2019Multilingual} take advantage of the similarity between the corpus of each language. During the multilingual model training, they start training with uniform-sampling from each corpus at first, and then gradually increase the training samples from more related corpora. \cite{im} proposed a stacked architecture where the first network is a BN feature extractor and the second network is the acoustic model. \cite{Hardik2020Multilingual} used MTL-SOL (Multi-Task Learning Structured Output Layer) framework by conducting language-specific phoneme recognition and multilingual acoustic modeling together. \cite{Liu2020MultilingualGH} demonstrated a multilingual grapheme-based ASR model learned on seven different languages and used multiple data augmentation alternatives within languages to further leverage the complementarity with multilingual modeling.

Meta-learning has also been applied to low-resource ASR. Meta learning is used to solve the problem of fast adaption of unseen data, which corresponds to the setting of low-resource ASR. MetaASR~\cite{hsu2020meta} meta-learns initialization parameters from pre-training tasks in different languages, quickly adapting to unseen target languages. Winata \textit{et al.}~\cite{winata2020meta} proposed a meta-transfer learning method that explores the transfer of knowledge from monolingual rich-resource language to low-resource languages. There are also some methods of learning the optimal network structure for multilingual ASR automatically. An efficient gradient-based architecture search algorithm is applied in DARTS-ASR, avoiding to design the model architecture and hyperparameters manually~\cite{Chen2020}. Language Adaptive DNNs train another neural network to encode language specific features, and add this language information to the input features of the network~\cite{Muller2016Language}. \cite{Xiao2021AdversarialMS} developed an adversarial meta sampling (AMS) approach to improve multilingual meta-learning ASR (MML-ASR). AMS calculates adaptively the task sampling probability for each source language when sampling tasks in MML-ASR. The AMS framework excellently tackle the task-imbalance problem caused by language tasks difficulties and quantities by well-designed sampling approach.

\section{Methodology}

\subsection{Speech Recognition Architecture}
The end-to-end CNN-RNN-DNN model has made great success in the speech recognition task on large corpus of English and Mandarin~\cite{ds2}. In this work, we used this CNN-RNN-DNN network for low resource ASR. 

\begin{figure}[ht]
	\begin{center}
		\centerline{\includegraphics[width=0.72\columnwidth]{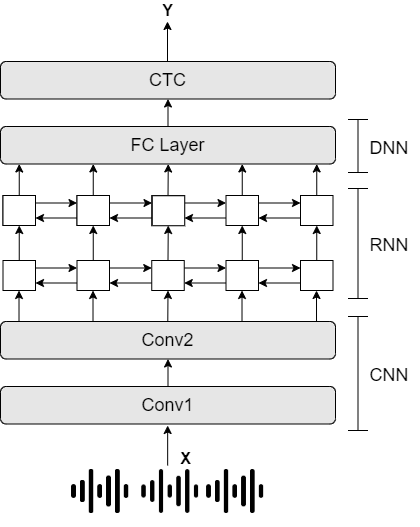}}
		\caption{The architecture of CNN-RNN-DNN network for low resource speech recognition task. The bottom convolutional layers (CNN) extract the local features of the audio. The middle recurrent layers (RNN) model the long-time dependency of the feature sequence. The upper deep neural layers (DNN) map the features to the vocabulary of target language. The whole network is trained by the CTC loss between the output hidden states and gold text sequence.}
		\label{fig1}
	\end{center}
\end{figure} 

As depicted in Fig~\ref{fig1}, the network contains three components: (1) convolutional layers, extracting local-time features, (2) recurrent layers, modeling long-time dependency, (3) deep neural layers, projecting the hidden states to the vocabulary. The network is trained via the CTC loss. For a special language $l$, the low resource speech recognition network maps the input acoustic feature sequence $X^l=(x_0^l, x_1^l, ..., x_T^l)$ to the text sequence $Y^l=(y_0^l, y_1^l, ..., y_U^l)$. In which, $T$ is the frame length of the audio, and $U$ is the text length of the sentence. The architecture is several convolutional layers, followed by several recurrent layers, and lastly fully connected layers.

For convolutional layers, the hidden states $O_n^l$ of layer $n$ and language $l$ are computed via the convolutions over the bottom layer output $O_{n-1}^l$.
\begin{equation} 
O_{n}^l=F_n(W_nO_{n-1}^l)
\end{equation}
In which, $W_n$ is the convolutional weight and $F_n(\cdot)$ is the activation function. $O_0^l$ represents the input audio features $X^l$. Following the convolutional layers are the bidirectional recurrent layers with the forward features $\overrightarrow O_{n}^l$ and backward features $\overleftarrow O_{n}^l$. The output $O_{n,t}^l$ of recurrent layer at time step $t$ is computed as:
\begin{equation} 
\overrightarrow O_{n,t}^l=G_n(O_{n-1,t}^l, \overrightarrow O_{n,t-1}^l)
\end{equation}
\begin{equation} 
\overleftarrow O_{n,t}^l=G_n(O_{n-1,t}^l, \overleftarrow O_{n,t+1}^l)
\end{equation}
\begin{equation} 
O_{n,t}^l=\overrightarrow O_{n,t}^l+\overleftarrow O_{n,t}^l
\end{equation}
In which, $G_n(\cdot)$ is the recurrent operations:
\begin{equation} 
\overrightarrow O_{n,t}^l=F_n(W_nO_{n-1,t}^l+\overrightarrow U_n\overrightarrow O_{n,t-1}^l+b_n)
\end{equation}
where $W_n$ is the weights of input features, $\overrightarrow U_n$ is the weights of recurrent features, and $b_n$ is the bias item. In our works, the $G_n(\cdot)$ could be replaced with more complex operations, gated recurrent units (GRU). After convolutional and recurrent layers, several fully connected layers are computed as:
\begin{equation} 
O_{n}^l=F_n(W_nO_{n-1}^l+b_n)
\end{equation}

CTC loss is applied to the output hidden feature map $O_{N}^l$ of output layer $N$. CTC predicts a frame-level alignment between the input sequence $X^l$ and the output sequence $Y^l$. The CTC loss could be defined as follows:
\begin{equation}
\mathcal{L}_{ctc}^l \triangleq -\ln~P_{ctc}(Y^l|O_{N}^l) = -\ln~\prod_{t}P(y^l|O_{N,t}^l)
\end{equation}
Where $P(y^l|O_{N,t}^l)$ is the probability of the corresponding text at frame step $t$ of the network output. 


\begin{figure*}[ht]
 \centering
 \subfigure[Traditional Transfer Learning] { \label{fig2a}
  \includegraphics[width=0.4\columnwidth]{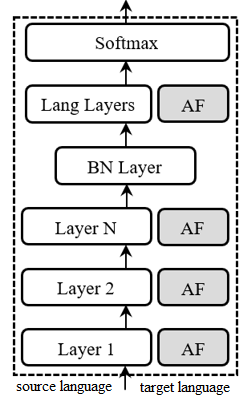}
 }
 \subfigure[Traditional Multilingual Structure] { \label{fig2b}
  \includegraphics[width=0.48\columnwidth]{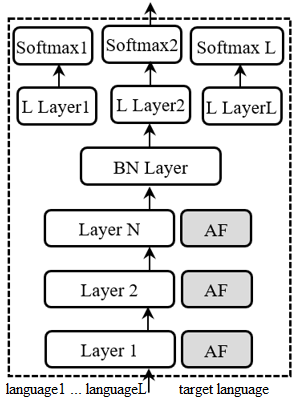}
 }
 \subfigure[The Proposed AANET Architecture] { \label{fig2c}
  \includegraphics[width=0.72\columnwidth]{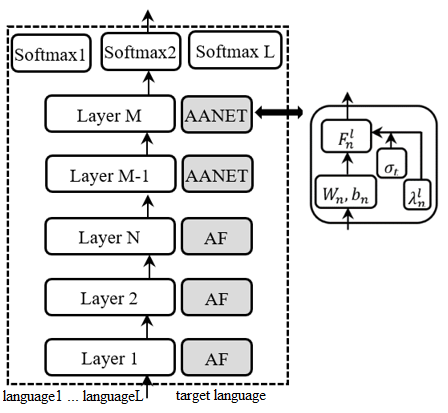}
 }
 \caption{The traditional transfer learning (a) and multilingual (b) structures are training the models in source large or multilingual corpus, to produce the bottleneck (BN) layer features. And then, the upper layers are finetuned in the target corpus for target ASR task. By contrast, the proposed Adaptive Activation Network (AANET) architecture (c) only replaces the Activation Function (AF) of the upper layers, and uses these AANET to model the relevance and difference among different languages.}
 \label{pig-full}
\end{figure*}

\subsection{Adaptive Activation Network}

Activation function introduces the nonlinear characteristics to the model, and plays an important role in the network of speech recognition. The network layer usually has three components, activation function, weight and bias parameters. In this work, we introduced the adaptive activation network~\cite{2020Adaptive}, replacing the traditional activation functions (like sigmoid, tanh, ReLu, etc.). 
For low resource ASR tasks, we applied the activation function $F_n^l$ to the target language $l$, and shared weight $W_n$ and bias $b_n$ across different languages.
\begin{equation} 
O_{n}^l=F_n^l(W_nO_{n-1}^l + b_n)
\end{equation}
In which, $W_n$ and $b_n$ were the shared weight and bias parameters. $F_n^l$ denotes the specific activation function of different language $l$ at $n$-th layer. The defination of $F_n^l$ is a set of basis functions as follows:
\begin{equation} 
F_n^l(\cdot) = \sum_{i=1}^{M}\lambda_{n}^{l}(i)\sigma_{i}(\cdot)
\end{equation}
where $\{\sigma_{i}\}_{i=1}^{M}$ represents $M$ different basis activation function, and $\lambda_{n}^{l} = [\lambda_{n}^{l}(1), ..., \lambda_{n}^{l}(M)]\in \mathbb{R}^M$ are the coordinates of bases $\{\sigma_{i}\}_{i=1}^{M}$. In this work, we also chose the adaptive piecewise linear (APL) activation units\cite{2020Adaptive}, to parameterize $F_n^l(\cdot)$. The definition of $F_n^l(\cdot)$ is as following:
\begin{equation} 
F_n^l(x) =  max(0, x) + \sum_{i=1}^{M}\lambda_{n}^{l}(i)max(0, -x + b_i)
\end{equation}
where $b_i$ is also the trainable parameter, and $M$ denotes the number of different APL units. It should be noted that the first item of $F_n^l(x)$ is the traditional ReLu activation function, and this mechanism enables that the adaptive activation network could have equivalent or better performance than existing activation functions.

\subsection{Cross-lingual Learning}
We denote the Adaptive Activation Network as ``AANET'', while the traditional Activation Functions are denoted as ``AF''. As depicted in Fig~\ref{fig2a}, the traditional transfer learning method is: (1) pre-training the network in source large corpus, (2) fixing the bottom layers to extract bottleneck (BN) features, and (3) fine-tuning the upper layers for the target low resource language. By contrast, our method leveraged the adaptive activation network to replace activation function of upper layers. Thus, the information in weight matrix and bias could be reserved.
\begin{equation}
\mathcal{L}_{pre} = \mathcal{L}_{ctc}^{l_0}
\end{equation}

More specifically, our cross-lingual learning method is: (1) pre-training the model, which contains adaptive activation network $F_n^{l_0}$, in a large source corpus $l_0$ by the source CTC loss $\mathcal{L}_{ctc}^{l_0}$, (2) applying a new adaptive activation network $F_n^{l_1}$ to the upper layers, and maintaining the weight and bias paramters, (3) fine-tuning these new adaptive activations for the target language $l_1$ by the target CTC loss $\mathcal{L}_{ctc}^{l_1}$. It should be noted that only upper layers' activation functions are replaced. Because bottom layers are leveraged to extract speech feature, and could not distinguish different unique features among different languages.
\begin{equation}
\mathcal{L}_{fine} = \mathcal{L}_{ctc}^{l_1}
\end{equation}

\subsection{Multilingual Learning}

Besides the cross-lingual learning, another solution for low resource ASR is multilingual architecture. As depicted in Fig~\ref{fig2b}, traditional multilingual approach is: (1) multilingual model shares the bottom layers until bottleneck layer in a single network, (2) every target language has a unique branch of upper layers and softmax layer, and (3) the whole network is jointly trained through a multi-task learning loss of each language branch.

Our multilingual learning method is shown in Fig~\ref{fig2c}. Several target languages $(l_1, l_2, ..., l_L)$ could be trained in a single network, with different adaptive activation networks of each language. The loss function of multilingual training contains two parts. The first item is the CTC loss of each language. The second item is the loss of distinguishing different languages in the multilingual process.
\begin{equation}
\mathcal{L}_{multi} = \sum_{i=1}^{L}\mathcal{L}_{ctc}^{l_i}+\alpha \mathcal{L}_{mtl}
\end{equation}

We introduced the trace-norm function to reflect the relevance of different languages. The definition of multi-task languages loss $\mathcal{L}_{mtl}$ is as following:
\begin{equation} 
\begin{aligned}
\mathcal{L}_{mtl}  = trace(\sqrt{\lambda_n {\lambda_n}^T})
\end{aligned}
\end{equation}
where $\sqrt{\cdot}$ denotes the square root of matrix, $\lambda_n=[\lambda_n^1,...,\lambda_n^L]\in \mathbb{R}^{L\times M}$ represents the activation coefficient matrix of layer $n$. $L$ is the number of languages. The trace-norm $trace(\cdot)$ is proven to be the convex envelope of the matrix rank~\cite{convex}. By minimizing the multi-task language loss $\mathcal{L}_{mtl}$, we aim to find a coefficient matrix $\lambda_n$ with lowest rank~\cite{huang2016lowrank}. Since the rank of a matrix equals the maximum number of linearly independent column vectors, $\lambda_n$ will have as many linear dependent column vectors as possible if its rank is minimized.  
Thus, it leads to a higher correlation between the languages in multilingual learning by minimizing the trace-norm term, which corresponds to our initial goal that the knowledge sharing between multiple tasks should be encouraged in multi-task learning. The low-rank assumption of the model parameters is often considered in previous multi-task learning work~\cite{Argyriou07convexmulti-task, Kei10tracenorm}. Our proposed regularization term $\mathcal{L}_{mtl}$ is aligned with the hypothesis on the assumption and linear dependency of coefficient matrix.                    

In addition, we could combine the cross-lingual learning and multilingual learning together, to further improve the model performance. It should be noted that multiple source languages can be used in the cross-lingual learning stage with the multilingual loss $\mathcal{L}_{multi}$, instead of monolingual CTC loss.

\section{Visualization}
\begin{figure}[h]
	\begin{center}
		\centerline{\includegraphics[width=0.9\columnwidth]{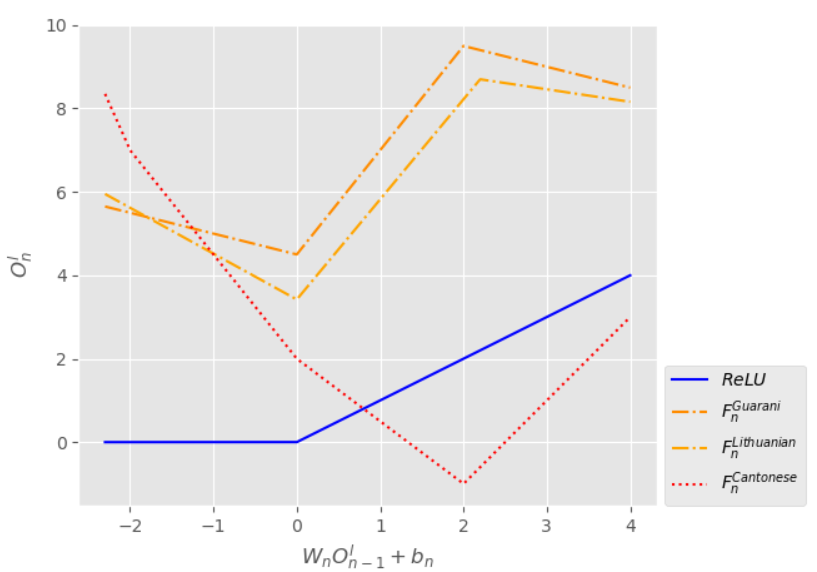}}
		\caption{The tradition ReLU activation function (blue) and the adaptive activation networks of Guarani, Lithuanian and Cantonese (orange, dark orange and red) in layer $n$. The adaptive activation networks of Guarani and Lithuanian are more similar, and differ from that of Cantonese a lot. Adaptive activation network introduces more non-linearity into the neural network and increases the learning capability of ASR system.}
		\label{relu_and_aanet}
	\end{center}
\end{figure}

In order to further illustrate the benefits of proposed adaptive activation network, we display an example about the differences between the traditional ReLU activation function and adaptive activation network in Fig~\ref{relu_and_aanet}. The solid blue line represents the traditional ReLU activation function, which has no parameters to learn during training. The dash-dotted orange and dark orange line denotes adaptive activation network learned from Guarani and Lithuanian languages, respectively. The dotted red line indicates adaptive activation network learned from Cantonese. The horizontal axis is the input of activation function $W_nO_{n-1}^l + b_n$, which is the output of layer $n-1$ for language $l$ operated by weight and bias in layer $n$; the vertical axis is the output of activation function $O_{n}^l$. As Guarani and Lithuanian share more common features, the adaptive activation networks learned from them are more similar. Cantonese belongs to another language family, thus, its adaptive activation network differs a lot. 

It is clear that, adaptive activation network introduces more non-linearity into the neural network. The neural network is able to learn more complicated relationship between hidden layers. Thus, from the mathematical point of view, the proposed adaptive activation network increases the learning capability of our ASR system. From the perspective of application, adaptive activation network can fully leverage the correlation between languages. As described previously, the weights and bias are shared in TAAN. Thus, similar languages, like Guarani and Lithuanian, can be expressed using similar adaptive activation function, instead of learning this similarity through multiple hidden layers with traditional ReLU activation. 

\section{Experiments}
In this section, we evaluated the adaptive activation approach in low resource ASR tasks. We investigated two kinds of CNN-RNN-DNN networks with different model sizes. The experiment results of proposed cross-lingual learning and multilingual learning approaches were obtained, and compared with from-scratch training and bottleneck features methods. We also showed the performance improvement by combining the cross-lingual and multilingual learning, and explored different configurations of adaptive activation layers.

\subsection{Datasets}
We used IARPA Babel datasets~\cite{Gales2014Gales} to conduct the experiments of proposed adaptive activation method for low resource ASR. The IARPA Babel datasets consist of conversational telephone speech for $28$ languages collected across a variety of environments. $5$ languages (Amharic, Cantonese, Guarani, Igbo, Lithuanian) are applied in our experiments. $3$ languages (Guarani, Igbo, Lithuanian) are used as the source languages, and $2$ languages (Amharic, Cantonese) are used as the target languages. For two target languages, we randomly split each language dataset into two parts. Three-fourths of the dataset is used for training, and the other one-fourths is used for testing. We demonstrate the detail information of used datasets in Table~\ref{tab1}.

\begin{table}[h]
	\caption{The Used datasets for experiments}
	\centering
	\label{tab1}
	\begin{tabular}{lll}
		\toprule
		\textbf{Source Languages} & \textbf{Length, Hours (h)} & \textbf{Speakers' Ages (years)}\\
		Guarani & 198 & 16-67\\
		Igbo & 207 & 16-67\\
		Lithuanian & 210 & 16-71\\
		\midrule
		\textbf{Target Languages} & \textbf{Length, Hours (h)} & \textbf{Speakers' Ages (years)}\\
		Amharic & 204 & 16-60\\
		Cantonese & 215 & 16-67\\
		\bottomrule
	\end{tabular}
\end{table}

\subsection{Setups}
We designed two kinds of CNN-RNN-DNN networks with different layers, which is called CRD-Small and CRD-Large. The hyperparameters of CRD-Small and CRD-Large are listed in Table~\ref{tab2}. For CRD-Small model, the adaptive activation is applied in the first DNN layer and last GRU layer. For CRD-Large model, the first DNN layer and both of last two GRU layers use the adaptive activation.

\begin{table}[h]
	\caption{The Hyperparameters of CRD-Small and CRD-Large models}
	\centering
	\label{tab2}
	\scalebox{0.9}{
	\begin{tabular}{llllll}
		\toprule
		\textbf{Model}&\textbf{CNN} &\textbf{RNN} & \textbf{DNN} & \textbf{AANET}\\
		\midrule
		&2 Conv layers &2 GRU layers &2 FC layers & \multirow{3}{*}{1GRU,1DNN} \\
		CRD-Small &$5\times 5$ filters & &\\
		&32 units  &128 units &1024 units\\
		\midrule
		&3 Conv layers &3 GRU layers &2 FC layers & \multirow{3}{*}{2GRU,1DNN} \\
		CRD-Large  &$5\times 5$ filters & &\\
		&64 units  &256 units &1024 units\\
		\bottomrule
	\end{tabular}}
\end{table} 

\begin{table}[h]
	\centering
	\caption{The Hyperparameters of Experimental Setups}
	\label{tab3}
	\begin{tabular}{ll}
		\toprule
		\textbf{Unit Name} & \textbf{Hyperparameters}\\
		\midrule
		\multirow{2}{*}{Speech Features Extraction} & sliding window = 25 ms\\
		& frame-shift = 10 ms\\
		\midrule
		Speech Frame Representation & 40-dimensional log Mel-filter bank\\
		\midrule
		\multirow{3}{*}{Adam Optimizer} & learning rate = 0.001\\
		& $\beta_1$ = 0.9\\
		& $\beta_2$ = 0.98\\
		\midrule
		Gradient Cropping Coefficient & [-1,1]\\
		\midrule
		Beam Width & 10\\
		\bottomrule
	\end{tabular}
\end{table}

\begin{table*} [ht]
	\centering 
	\caption{Results of Different Training Strategies with Adaptive Activation Network, WER (\%)}
	\label{tab4}
	\begin{tabular}{p{4.2cm}p{4.2cm}p{4.2cm}p{1.5cm}p{1.5cm}} 
		\toprule
		\textbf{Model} & \textbf{Pre-training Data} & \textbf{Fine-tuning Data} & \textbf{Amharic} & \textbf{Cantonese}\\  
		\midrule
		CRD-Small + FS & - & Amharic, Cantonese & 71.2 & 58.3\\
		CRD-Small + BN & Guarani, Igbo, Lithuanian & Amharic, Cantonese & 69.1 & 55.1\\      
		CRD-Small + CL& Guarani, Igbo, Lithuanian & Amharic, Cantonese & 68.2 & 53.2 \\ 
		CRD-Small + ML & - & Guarani, Igbo, Lithuanian, Amharic, Cantonese & 68.9 & 56.2\\
		CRD-Small + CL \& ML & Guarani, Igbo, Lithuanian & Guarani, Igbo, Lithuanian, Amharic, Cantonese & \textbf{67.3} & \textbf{52.9} \\
		\midrule
		CRD-Large + FS & - & Amharic, Cantonese & 68.9 & 57.7\\
		CRD-Large + BN & Guarani, Igbo, Lithuanian & Amharic, Cantonese & 66.3 & 54.6\\      
		CRD-Large + CL & Guarani, Igbo, Lithuanian & Amharic, Cantonese & 66.2 & 51.3\\ 
		CRD-Large + ML & - & Guarani, Igbo, Lithuanian, Amharic, Cantonese & 67.8 & 54.1\\
		CRD-Large + CL \& ML & Guarani, Igbo, Lithuanian & Guarani, Igbo, Lithuanian, Amharic, Cantonese & \textbf{66.3} & \textbf{51.1} \\
		\bottomrule
	\end{tabular}  
\end{table*}

Besides different model size, we also compared our models with two other methods, from-scratch training and bottleneck features.
\begin{itemize}
	\setlength{\itemsep}{0pt}
	\setlength{\parsep}{0pt}
	\setlength{\parskip}{0pt}
	\item \textbf{From-Scratch Training}: Directly train the model using the training set of target language without pre-training or multi-task learning.
	\item \textbf{Bottleneck Features}~\cite{icasp1}: Insert a Bottleneck (BN) layer between the first and second DNN layer, and the dimension of BN layer is set to $80$.
\end{itemize}

The speech features were extracted with a $25$-ms sliding window with a $10$-ms frame shift. Each speech frame was represented by $40$-dimensional log Mel-filter bank (Fbank) features. All of our experiments were implemented by TensorFlow2. The Adam optimizer was used to minimize the loss function. The learning rate was set to $0.001$, and the $\beta_1$ and $\beta_2$ were $0.9$ and $0.98$. During the model training, the gradient cropping strategy was also applied, and the gradient of each parameter was limited between $-1$ and $1$. During the inference stage, we used beam search to obtain the recognition results, and the beam width was set to $10$. The hyperparameters of our experimental setups are shown in Table~\ref{tab3}. For fair comparison, all of the models were evaluated by word error rate (WER), and not any language model was leveraged in the decoding.

\subsection{Results}

The results of different training methods are shown in Table~\ref{tab4}. We denote different training strategies as following:
\begin{itemize}
	\setlength{\itemsep}{0pt}
	\setlength{\parsep}{0pt}
	\setlength{\parskip}{0pt}
	\item \textbf{FS}: From-Scratch Training
	\item \textbf{BN}: Bottleneck Features
	\item \textbf{CL}: Cross-lingual Learning
	\item \textbf{ML}: Multilingual Learning
\end{itemize}

We can observe that our cross-lingual learning approach with adaptive activation outperforms the from-scratch training and traditional bottleneck feature approaches. It confirmed that the adaptive activation method is effective for low resource ASR task. In addition, combining the cross-lingual learning and multilingual learning will further improve the performance, and achieve the best WER results. The combined CRD-Large + CL \& ML model achieves $3.5\%$ WER reduction than traditional CRD-Large + BN model on Cantonese language dataset. We could also observe that the larger CRD-Large model could have better performance than CRD-Small model. The combined CRD-Large + CL \& ML model has $1.0\%$ WER improvement on Amharic dataset and $1.8\%$ on Amharic dataset, than CRD-Small + CL \& ML model.

In order to explore the function of adaptive activation, different GRU layers with adaptive activation networks are further investigated. We use CRD-Large model to compare different structures. The results are shown in Table~\ref{tab5}. We use AANET to replace different configurations of layers.  It can be inferred that increasing the number of adaptive activation layers could improve the model performance, but more layers might have limited improvement. Therefore, we chose $2$GRU,$1$DNN as the standard configuration of CRD-Large model.

\begin{table}[h]
	\caption{Different Configuration of Adaptive Activation Network, WER (\%)}
	\centering
	\label{tab5}
	\begin{tabular}{llll}
		\toprule
		\textbf{Model} & \textbf{AANET} & \textbf{Amharic} & \textbf{Cantonese}\\
		\midrule
		CRD-Large + CL & 1GRU,1DNN & 67.1 & 52.6 \\
		CRD-Large + CL & 2GRU,1DNN & 66.2 & \textbf{51.3} \\
		CRD-Large + CL & 3GRU,1DNN & \textbf{66.0} & 51.6 \\
		\bottomrule
	\end{tabular}
\end{table} 

\section{Conclusion}
This paper introduced adaptive activation network to the low resource multilingual speech recognition. Two kinds of end-to-end CNN-RNN-DNN networks are used for ASR tasks. We found that CRD-Large model with larger model size could have better performance than smaller CRD-Small model. The adaptive activations are applied to different languages, instead of fine-tuning the parameters of upper layers. We proposed cross-lingual learning and multilingual learning approaches to realize this adaptive strategy. The experiment results show that our approaches achieve better results than traditional bottleneck features method. Moreover, combining cross-lingual learning and multilingual learning could further improve the model performance. We also explored the configuration of adaptive activation layers, inferring that more adaptive activation layers lead to better WER results. The future works involved the exploration of better structure for multilingual ASR, such as hybrid CTC/attention model. We are also interested in investigating different strategies to measure the relevance of multi-languages, such as cosine similarity, euclidean distance, etc.

\section{Acknowledgement}
\label{sec:ack}
This paper is supported by the Key Research and Development Program of Guangdong Province under grant No.
2021B0101400003. Corresponding author is Jianzong Wang from Ping An Technology (Shenzhen) Co., Ltd (jzwang@188.com).

\clearpage
\bibliographystyle{IEEEbib}
\bibliography{IJCNN2022_AANET}

\begin{thebibliography}{10}

\bibitem{joint1}
Suyoun Kim, Takaaki Hori, and Shinji Watanabe,
\newblock ``Joint ctc-attention based end-to-end speech recognition using
  multi-task learning,''
\newblock in {\em IEEE International Conference on Acoustics, Speech and Signal
  Processing (ICASSP)}, 2017.

\bibitem{Hybrid2}
Haoran Miao, Gaofeng Cheng, Pengyuan Zhang, Ta~Li, and Yonghong Yan,
\newblock ``Online hybrid ctc/attention architecture for end-to-end speech
  recognition.,''
\newblock in {\em IEEE Conference of the International Speech Communication
  Association (INTERSPEECH)}, 2019.

\bibitem{Luo2021Multi}
Jian Luo, Jianzong Wang, Ning Cheng, Guilin Jiang, and Jing Xiao,
\newblock ``Multi-quartznet: Multi-resolution convolution for speech
  recognition with multi-layer feature fusion,''
\newblock in {\em IEEE Spoken Language Technology Workshop (SLT)}, 2021.

\bibitem{am}
Martin Karafiát, Baskar~Murali Karthick, Watanabe Shinji, Hori Takaaki, and
  Jan Černocký,
\newblock ``Analysis of multilingual sequence-to-sequence speech recognition
  systems,''
\newblock in {\em IEEE Conference of the International Speech Communication
  Association (INTERSPEECH)}, 2019.

\bibitem{2017An}
Sibo Tong, Philip~N. Garner, and Hervé Bourlard,
\newblock ``An investigation of deep neural networks for multilingual speech
  recognition training and adaptation,''
\newblock in {\em IEEE Conference of the International Speech Communication
  Association (INTERSPEECH)}, 2017.

\bibitem{icassp20201}
C.~{Du} and K.~{Yu},
\newblock ``Speaker augmentation for low resource speech recognition,''
\newblock in {\em IEEE International Conference on Acoustics, Speech and Signal
  Processing (ICASSP)}, 2020.

\bibitem{icassp20191}
Sibo Tong, Philip~N Garner, and Herv{\'e} Bourlard,
\newblock ``An investigation of multilingual asr using end-to-end lf-mmi,''
\newblock in {\em IEEE International Conference on Acoustics, Speech and Signal
  Processing (ICASSP)}, 2019.

\bibitem{Jia2020Large}
Xueli Jia, Jianzong Wang, Zhiyong Zhang, Ning Cheng, and Jing Xiao,
\newblock ``Large-scale transfer learning for low-resource spoken language
  understanding,''
\newblock in {\em IEEE Conference of the International Speech Communication
  Association (INTERSPEECH)}, 2020.

\bibitem{icassp20202}
J.~{Hsu}, Y.~{Chen}, and H.~{Lee},
\newblock ``Meta learning for end-to-end low-resource speech recognition,''
\newblock in {\em IEEE International Conference on Acoustics, Speech and Signal
  Processing (ICASSP)}, 2020.

\bibitem{Luo2021Cross}
Jian Luo, Jianzong Wang, Ning Cheng, Edward Xiao, Jing Xiao, Georg Kucsko,
  Patrick O’Neill, Jagadeesh Balam, Slyne Deng, Adriana Flores, Boris
  Ginsburg, Jocelyn Huang, Oleksii Kuchaiev, Vitaly Lavrukhin, and Jason Li,
\newblock ``Cross-language transfer learning and domain adaptation for
  end-to-end automatic speech recognition,''
\newblock in {\em IEEE International Conference on Multimedia and Expo (ICME)},
  2021.

\bibitem{alter}
William Hartmann, Roger Hsiao, and Stavros Tsakalidis,
\newblock ``Alternative networks for monolingual bottleneck features,''
\newblock in {\em IEEE International Conference on Acoustics, Speech and Signal
  Processing (ICASSP)}, 2017.

\bibitem{ACL1}
Julius Kunze, Louis Kirsch, Ilia Kurenkov, Andreas Krug, Jens Johannsmeier, and
  Sebastian Stober,
\newblock ``Transfer learning for speech recognition on a budget,''
\newblock in {\em IEEE Proceedings of the 58th Annual Meeting of the
  Association for Computational Linguistics (ACL)}, 2017.

\bibitem{ds2}
Dario Amodei, Sundaram Ananthanarayanan, Rishita Anubhai, Jingliang Bai, and
  Zhenyao Zhu,
\newblock ``Deep speech 2: End-to-end speech recognition in english and
  mandarin,''
\newblock in {\em IEEE International Conference on Machine Learning (ICML)},
  2015.

\bibitem{icasspx}
Tom Sercu, Christian Puhrsch, Brian Kingsbury, and Yann LeCun,
\newblock ``Very deep multilingual convolutional neural networks for lvcsr,''
\newblock in {\em IEEE International Conference on Acoustics, Speech and Signal
  Processing (ICASSP)}, 2016.

\bibitem{2016Multilingual}
S~Thomas, K~Audhkhasi, J~Cui, B~Kingsbury, and B~Ramabhadran,
\newblock ``Multilingual data selection for low resource speech recognition,''
\newblock in {\em IEEE Conference of the International Speech Communication
  Association (INTERSPEECH)}, 2016.

\bibitem{2019Multilingual}
Xinjian Li, Siddharth Dalmia, Alan~W Black, and Florian Metze,
\newblock ``Multilingual speech recognition with corpus relatedness sampling,''
\newblock in {\em IEEE Conference of the International Speech Communication
  Association (INTERSPEECH)}, 2019.

\bibitem{im}
Tanel Alumäe, Stavros Tsakalidis, and Richard Schwartz,
\newblock ``Improved multilingual training of stacked neural network acoustic
  models for low resource languages,''
\newblock in {\em IEEE Conference of the International Speech Communication
  Association (INTERSPEECH)}, 2016.

\bibitem{icasp1}
Jiangyan Yi, Jianhua Tao, and Ye~Bai,
\newblock ``Language-invariant bottleneck features from adversarial end-to-end
  acoustic models for low resource speech recognition,''
\newblock in {\em IEEE International Conference on Acoustics, Speech and Signal
  Processing (ICASSP)}, 2019.

\bibitem{yi2019Language}
Jiangyan Yi, Jianhua Tao, Zhengqi Wen, and Ye~Bai,
\newblock ``Language-adversarial transfer learning for low-resource speech
  recognition,''
\newblock {\em IEEE Transactions on Audio, Speech, and Language Processing},
  2019.

\bibitem{2020Adaptive}
Yingru Liu, Xuewen Yang, Dongliang Xie, Xin Wang, Li~Shen, Haozhi Huang, and
  Niranjan Balasubramanian,
\newblock ``Adaptive activation network and functional regularization for
  efficient and flexible deep multi-task learning,''
\newblock in {\em Association for the Advancement of Artificial Intelligence
  (AAAI)}, 2020.

\bibitem{Luo2021Dropout}
Jian Luo, Jianzong Wang, Ning Cheng, and Jing Xiao,
\newblock ``Dropout regularization for self-supervised learning of transformer
  encoder speech representation,''
\newblock in {\em IEEE Conference of the International Speech Communication
  Association (INTERSPEECH)}, 2021.

\bibitem{zhang2021xlst}
Zi-Qiang Zhang, Yan Song, Ming-Hui Wu, Xin Fang, and Li-Rong Dai,
\newblock ``Xlst: Cross-lingual self-training to learn multilingual
  representation for low resource speech recognition,''
\newblock in {\em arXiv:2103.08207 2021}, 2019.

\bibitem{Kannan2019Large}
Anjuli Kannan, Arindrima Datta, Tara Sainath, Eugene Weinstein, Bhuvana
  Ramabhadran, Yonghui Wu, Ankur Bapna, Zhifeng Chen, and Seungji Lee,
\newblock ``Large-scale multilingual speech recognition with a streaming
  end-to-end model,''
\newblock in {\em IEEE Conference of the International Speech Communication
  Association (INTERSPEECH)}, 2019.

\bibitem{Kumar2021AnEO}
Ankith Jain~Rakesh Kumar and Rajesh~Kumar Aggarwal,
\newblock ``An exploration of semi-supervised and language-adversarial transfer
  learning using hybrid acoustic model for hindi speech recognition,''
\newblock {\em Journal of Research in International Education (JRIE)}, pp.
  1--16, 2021.

\bibitem{dalmia2018sequence}
Siddharth Dalmia, Ramon Sanabria, Florian Metze, and Alan~W Black,
\newblock ``Sequence-based multi-lingual low resource speech recognition,''
\newblock in {\em IEEE International Conference on Acoustics, Speech and Signal
  Processing (ICASSP)}, 2018.

\bibitem{hou2020large}
Wenxin Hou, Yue Dong, Bairong Zhuang, Longfei Yang, Jiatong Shi, and Takahiro
  Shinozaki,
\newblock ``Large-scale end-to-end multilingual speech recognition and language
  identification with multi-task learning.,''
\newblock in {\em IEEE Conference of the International Speech Communication
  Association (INTERSPEECH)}, 2020.

\bibitem{Hardik2020Multilingual}
Hardik~B. Sailor and Thomas Hain,
\newblock ``Multilingual speech recognition using language-specific phoneme
  recognition as auxiliary task for indian languages,''
\newblock in {\em IEEE Conference of the International Speech Communication
  Association (INTERSPEECH)}, 2020.

\bibitem{Liu2020MultilingualGH}
Chunxi Liu, Qiao chu Zhang, Xiaohui Zhang, Kritika Singh, Yatharth Saraf, and
  Geoffrey Zweig,
\newblock ``Multilingual graphemic hybrid asr with massive data augmentation,''
\newblock in {\em Workshop on Spoken Language Technologies for Under-resourced
  languages (SLTU)}, 2020.

\bibitem{hsu2020meta}
Jui-Yang Hsu, Yuan-Jui Chen, and Hung-yi Lee,
\newblock ``Meta learning for end-to-end low-resource speech recognition,''
\newblock in {\em IEEE International Conference on Acoustics, Speech and Signal
  Processing (ICASSP)}, 2020.

\bibitem{winata2020meta}
Genta~Indra Winata, Samuel Cahyawijaya, Zhaojiang Lin, Zihan Liu, Peng Xu, and
  Pascale~Ngan Fung,
\newblock ``Meta-transfer learning for code-switched speech recognition,''
\newblock in {\em IEEE Proceedings of the 58th Annual Meeting of the
  Association for Computational Linguistics (ACL)}, 2020.

\bibitem{Chen2020}
Yi-Chen Chen, Jui-Yang Hsu, Cheng-Kuang Lee, and Hung yi~Lee,
\newblock ``Darts-asr: Differentiable architecture search for multilingual
  speech recognition and adaptation,''
\newblock in {\em IEEE Conference of the International Speech Communication
  Association (INTERSPEECH)}, 2020.

\bibitem{Muller2016Language}
Markus Muller, Sebastian Stüker, and Alex Waibel,
\newblock ``Language adaptive dnns for improved low resource speech
  recognition,''
\newblock in {\em IEEE Conference of the International Speech Communication
  Association (INTERSPEECH)}, 2016.

\bibitem{Xiao2021AdversarialMS}
Yubei Xiao, Ke~Gong, Pan Zhou, Guolin Zheng, Xiaodan Liang, and Liang Lin,
\newblock ``Adversarial meta sampling for multilingual low-resource speech
  recognition,''
\newblock in {\em Association for the Advancement of Artificial Intelligence
  (AAAI)}, 2021.

\bibitem{convex}
Fredrik Andersson, Marcus Carlsson, and Carl Olsson,
\newblock ``Convex envelopes for fixed rank approximation,''
\newblock {\em Optimization Letters}, 2017.

\bibitem{huang2016lowrank}
Shimeng Huang and Henry Wolkowicz,
\newblock ``Low-rank matrix completion using nuclear norm with facial
  reduction,''
\newblock {\em The Journal of Gastrointestinal Oncology (JGO)}, 2016.

\bibitem{Argyriou07convexmulti-task}
Andreas Argyriou, Theodoros Evgeniou, and Massimiliano Pontil,
\newblock ``Convex multi-task feature learning,''
\newblock in {\em Machine Learning}, 2007.

\bibitem{Kei10tracenorm}
Ting Kei, Pong Paul, Tseng Shuiwang, and Ji~Jieping Ye,
\newblock ``Trace norm regularization: Reformulations, algorithms, and
  multi-task learning,''
\newblock {\em SIAM Journal on Optimization}, 2010.

\bibitem{Gales2014Gales}
M.J.F. Gales, Kate Knill, A.~Ragni, and Shakti Rath,
\newblock ``Speech recognition and keyword spotting for low-resource languages
  : Babel project research at cued,''
\newblock in {\em Workshop on Spoken Language Technologies for Under-resourced
  languages (SLTU)}, 2014.

\end{thebibliography}

\end{document}